\documentclass[twocolumn]{article}

\usepackage{amsmath,amssymb,amsfonts}
\usepackage{cite}
\usepackage{graphicx}
\usepackage{hyperref}
\usepackage{multirow}
\usepackage{setspace}
\usepackage{subfig} 
\usepackage{url}

\begin{document}

\title{The Shooting Regressor; Randomized Gradient-Based Ensembles}


\author{Nicholas T. Smith\\
\textit{\href{https://nicholastsmith.wordpress.com}{nicholastsmith.wordpress.com}}\\
\href{mailto:nicholastsmithblog@gmail.com}{nicholastsmithblog@gmail.com}}

\maketitle

\begin{abstract}
An ensemble method is introduced that utilizes randomization and loss function gradients to compute a prediction. Multiple weakly-correlated estimators approximate the gradient at randomly sampled points on the error surface and are aggregated into a final solution. A scaling parameter is described that controls a trade-off between ensemble correlation and precision. Numerical methods for estimating optimal values of the parameter are described. Empirical results are computed over a popular dataset. Inferential statistics on these results show that the method is capable of outperforming existing techniques in terms of increased accuracy.
\end{abstract}


\section{Introduction}
In a traditional regression scenario, a model is constructed which directly estimates a numerical output given some vector of input. This output is typically derived from the input via linear algebra or some algorithm which operates on the input vector. Many types of models fall into this broad description including but not limited to: linear regression, decision trees, random forest, support vector machines, and neural networks \cite{RandForest,GradBoost,CART,ESL}. In linear regression, for example, the output is derived from the input via a matrix multiplication using a vector whose entries are optimized over the training data.

The gradient boosting machine (GBM) described by Friedman comprises an approach to regression that is conceptually different than the one taken by the above models. Each of the above models directly estimates the predicted value. Instead, GBM indirectly estimates the predicted value using approximations of the gradient \cite{GBGradDesc}.

In GBM, an initial guess is refined using a sequence of models. The $i$-th model in the sequence attempts to estimate the gradient of the loss vector at the $i$-th step of a gradient descent process. Thus, the target value for the $i$-th model in the sequence is not the true target value, but the gradient of the loss function at the $i$-th step. To compute the final prediction, the sequential process of gradient descent is approximated using the models in sequence until the list of estimators is exhausted. It is in this way that GBM can be thought to indirectly compute the final prediction as opposed to a more direct approach such as random forest (RF).

\section{Background and Motivation}

Consider a GBM that minimizes mean-squared error (MSE). The loss function for such a model is defined to be

\begin{equation*}
	L(\hat{Y}) = (\hat{Y} - Y)^2 ,
\end{equation*}

where $\hat{Y}$ is the vector of predicted values and $Y$ is the vector of targets. The gradient of this function with respect to the predicted vector is simply the difference between the predicted and actual values. That is,

\begin{equation*}
	G_L(\hat{Y}) = \hat{Y} - Y .
\end{equation*}

From this, it is apparent that if one has a model capable of reliably estimating the gradient, then computation of the true target value at any stage is trivial. The negative of the gradient simply need be added to the current prediction to obtain a loss of zero. This can be seen algebraically as follows,

\begin{equation*}
\begin{gathered}
	L(I - G_L(I)) = (I-G_L(I) - Y)^2  \\
	= (I - (I - Y) - Y)^2 = 0 ,
\end{gathered}
\end{equation*}

where $I$ is the current prediction vector.

Thus, the problem of estimating the target value is directly reducible to the problem of estimating the gradient. From this it may be concluded that estimation of the gradient must be at least as difficult as estimation of the original target value.

This result is also intuitive geometrically. When defined in terms of the predicted values, the MSE loss function describes a high-dimensional paraboloid. Since paraboloids are convex functions, they only have a single global minimum and that value is nothing more than the vertex. Given a point on the surface of the paraboloid and the gradient, it is trivial to compute the vertex; the gradient is pointing directly at it.

\section{Methodology}

It is interesting to consider alternative methods which indirectly estimate the predicted values stemming from this geometric perspective. The approach taken here is to randomly sample the gradient at an arbitrary number $k$ of points around the global minimum. In this way, an arbitrary number of target vectors is produced that each point at the global minimum from a different angle.

\subsection{The Shooting Regressor}

Given a training matrix of $m$ samples with $n$ features, an $m \times 1$ vector of target values, and $k$ $m \times 1$ initial vectors, the gradient vector with respect to each initial point is computed using the formula

\begin{equation*}
	G_L(I_i)=I_i - Y .
\end{equation*}

For each of the $k$ points, the global minimum may be obtained by subtracting the gradient at the point from the point itself. Next, $k$ models are constructed where the $i$-th model attempts to approximate the gradient at the $i$-th sampling point around the global minimum. An ensemble is then constructed which simultaneously uses each of the $k$ estimators along with the $k$ initial points to obtain $k$ approximations of the global minimum which are aggregated into a final predicted value.

When the gradient can be estimated exactly, this process is redundant in that each of the $k$ estimators  produces an identical target value. The advantage of the approach is thus in smoothing out noise arising in estimating the gradient in real-world data. Essentially, this is accomplished by providing a diversity of potential solutions and then aggregating them together.

\begin{figure}
\center
\includegraphics[width=\columnwidth]{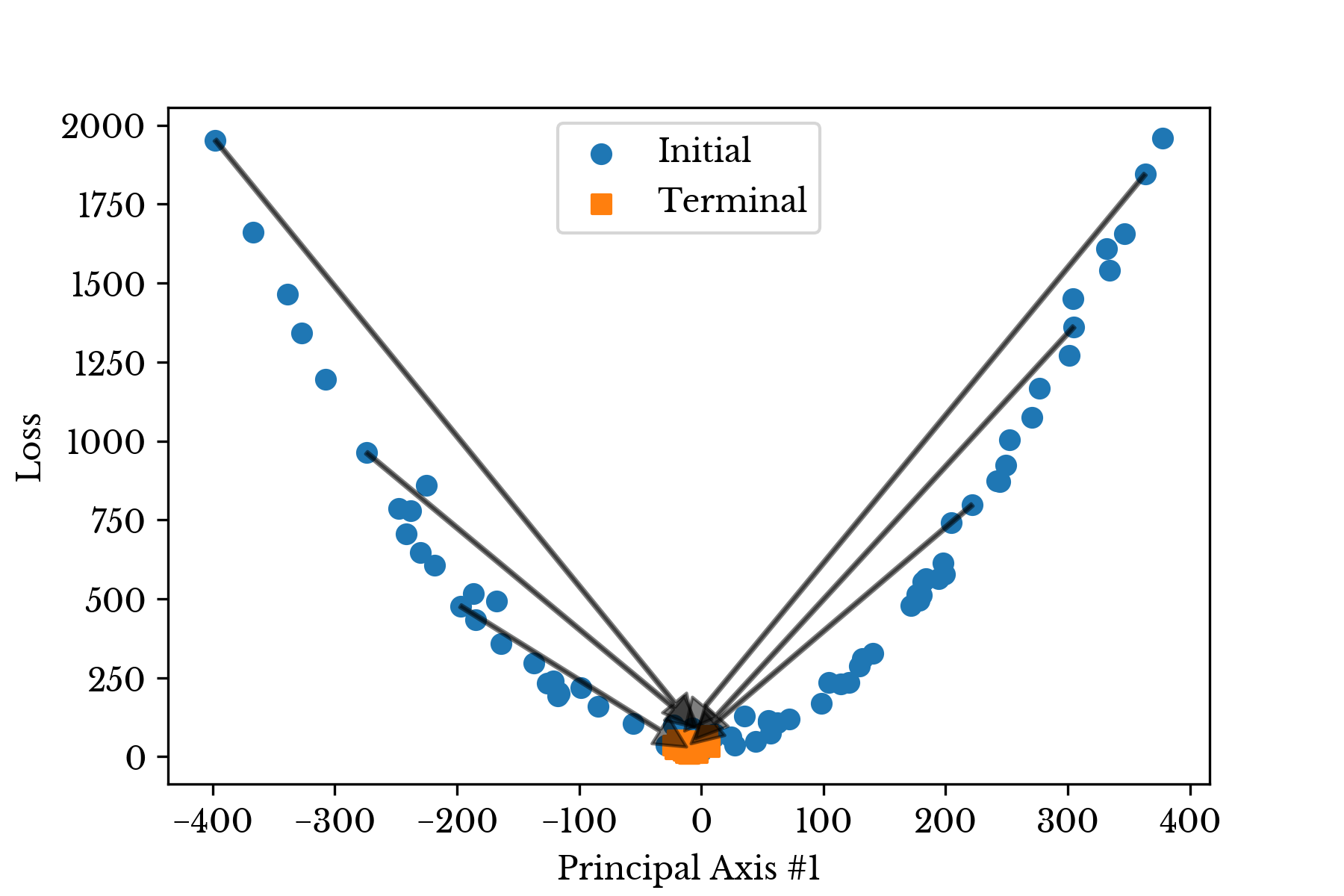}
\caption{Shooting Regressor PCA Projection}
\label{fig:ShootProj}
\end{figure}

Figure \ref{fig:ShootProj} describes the behavior of the shooting regressor. A synthetic dataset is constructed and multiple initial prediction functions are computed. The initial and terminal prediction vectors are projected into a 1-D subspace using principal component analysis (PCA) \cite{PCA}. Arrows connect the pairs of initial and terminal points representing the action of the gradient estimators.

The terminal points cluster around the target vector. The final estimate of the shooting regressor is the average of these terminal points.

The above approach describes a family of methods parameterized by $k$ functions for generating initial predictions and $k$ functions for estimating the gradient. A specific instance of this family is presented here using linear projections to generate the initial predictions and regression trees to estimate the gradient \cite{CART}.

\subsection{Selection of Initial Functions}

Consider a linear model constructed to minimize the MSE between the target and predicted values. The regression coefficients of such a model can be considered as random variables. It is known that these coefficients are distributed according to a multivariate normal distribution with mean $B$ and covariance 

\begin{equation*}
	\mathbf{C} = s^2(\mathbf{X}^T\mathbf{X})^{-1} ,
\end{equation*}

where $s^2$ is an approximation of the squared model error and $\mathbf{X}$ is the data matrix.

Intuitively, if different samples of data are collected from some underlying phenomenon, then different regression coefficients are likely to be obtained in each instance by the fitting procedure. In an effort to simulate this process, the initial value functions are sampled from this multivariate normal distribution. That is, each estimator is of the form

\begin{equation*}
\begin{gathered}
	M_i(\mathbf{X}) = \mathbf{X} (B+D_i) - \hat{G}_i(\mathbf{X}) \\
	                = \mathbf{X} I_i     - \hat{G}_i(\mathbf{X}) ,
\end{gathered}
\end{equation*}

where $I_i=B+D_i$, $D_i$ is distributed according to a multivariate normal distribution with mean $\mathbf{0}$ and covariance $\mathbf{C}$, and $\hat{G}_i(X)$ is the approximation of the gradient at $I_i$. Specifically,

\begin{equation*}
	\hat{G}_i(\mathbf{X}) \approx Y - \mathbf{X} (B + D_i) = Y - \mathbf{X} I_i ,
\end{equation*}

for $i=1,2, \ldots, k$. Thus, the final model output is

\begin{equation*}
	M(X) = \sum_{i}^{k}{(\mathbf{X} I_i - \hat{G}_i(\mathbf{X}))} / k .
\end{equation*}

Next, it is shown that such a strategy does not introduce bias into the estimation process. In order to focus on the effect of the initial value functions, assume that the second gradient term is itself an unbiased estimator of the gradient. Thus, if the initial value functions also produce an unbiased estimator, then the entire model itself is unbiased.

According to the assumptions of linear regression, the residuals of such a model are normally distributed with a mean of 0. Consider the output of the model as a random vector the value of which depends upon the randomly selected initial values. Consider the expected value of this model.

\begin{equation*}
\begin{gathered}
	E[M(X)] = E[\sum_{i}^{k}{\mathbf{X} I_i} / k]                                  \\
	        = E[\sum_{i}^{k}{(\mathbf{X} B + \mathbf{X} D_i)} / k]                 \\
	        = \sum_{i}^{k}{\mathbf{X} B} / k + E[\sum_{i}^{k}{\mathbf{X} D_i} / k] \\
	        = \sum_{i}^{k}{\mathbf{X} B} / k + \mathbf{X} E[\sum_{i}^{k}{D_i}] / k \\
	        = \sum_{i}^{k}{\mathbf{X} B} / k + \mathbf{X}\mathbf{0} / k  = \mathbf{X} B
\end{gathered}
\end{equation*}

Thus, if the estimation of the gradient is unbiased, then the entire model is unbiased. This follows from the fact that the linear model with coefficient vector $B$ is an unbiased estimator according to the assumptions of linear regression. In conclusion, this strategy for generating initial guesses does not introduce bias into the estimation process.

\subsection{A Scaling Parameter}

The covariance matrix of the above distribution may be scaled by some arbitrary constant. This constant $\nu$ controls the spread of the initial guesses around the best linear approximation and is a crucial component of the model. If $\nu$ is too small, the guesses are close together and the $k$ gradient vectors are highly correlated with each other. In such a situation, errors tend to accumulate instead of cancel and the ensemble offers little advantage over a single model.

As $\nu$ grows, the magnitude of the gradient vectors grows and their correlations with each other fall. In the limit as $\nu$ approaches infinity, the correlation between the gradients becomes the correlation between the initial guess offsets as the other terms are dominated by the scaled offset vector.

\begin{equation*}
\begin{gathered}
	\lim_{\nu \to \infty}{Corr(Y - M - \nu D_i, Y - M - \nu D_j)} \\
	= Corr(D_i, D_j)
\end{gathered}
\end{equation*}

If $\nu$ is too large, errors in estimation of the gradient are magnified and the resulting solution is poor. To model this trade-off, the norm of the correlation matrix between the gradient vectors is added to the norm of the scaled offset vectors. The goal is to find the value of $\nu$ that minimizes this objective function

\begin{equation}
\begin{gathered}
	\min_{\nu} {\left\lVert Corr(\mathbf{Z} - \nu \mathbf{I}) \right\rVert + \left\lVert \mathbf{Z} - \nu \mathbf{I} \right\rVert} ,
\end{gathered}
\label{ObjfExp}
\end{equation}

where $\mathbf{Z}$ is a matrix with $k$ columns that repeat the value $Y-M$ and $\mathbf{I}$ is a matrix with the $k$ initial vectors in its columns.

This expression is a polynomial in a single variable $\nu$ and can be found via numerical optimization. Expression \ref{ObjfExp} essentially describes a regularized solution for $\nu$ that minimizes the correlation in the gradient targets.

\begin{figure}
\center
\includegraphics[width=\columnwidth]{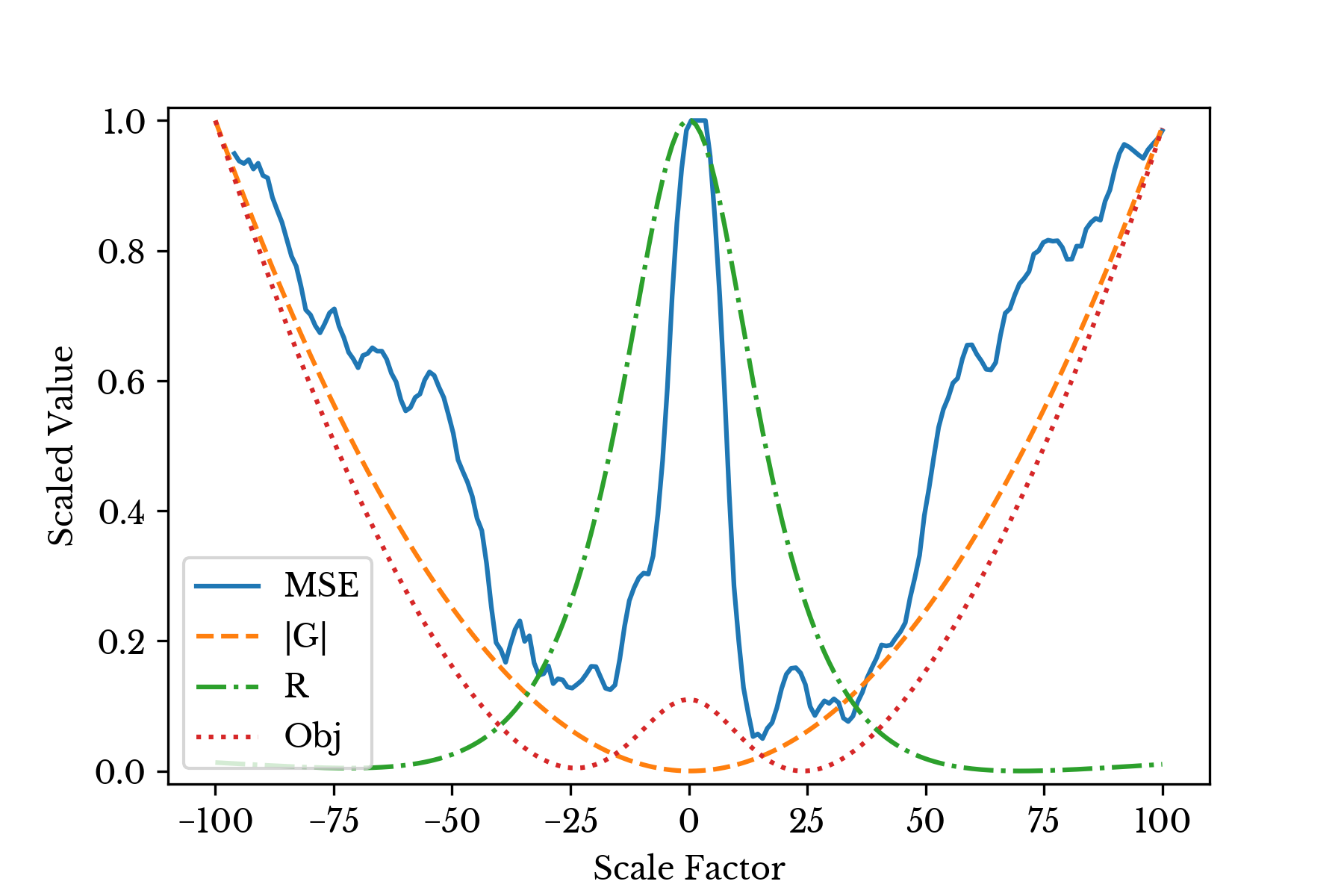}
\caption{Objective Function and Empirical Results}
\label{fig:NuObjf}
\end{figure}

Next, the behavior of $\nu$ is explored using the synthetic data referenced in Figure \ref{fig:ShootProj}. Figure \ref{fig:NuObjf} plots the curves for the correlation (R) and gradient magnitude ($|G|$) terms in green and orange respectively. Validation accuracy (MSE) for the model is computed over a range of $\nu$ values and is plotted in blue. As can be seen, the behavior of the model roughly follows the behavior of the objective function (Obj) depicted by the red curve.

Properties of the covariance operator are employed to efficiently calculate the correlation at each step in the minimization process.

\begin{equation*}
\begin{gathered}
	Corr(Z-\nu*Xi, Z-\nu*Xj)= \\
	\frac{(C_{zz} - \nu C_{zi} - \nu C_{zi} + \nu^2 C_{ij})}{\sqrt{(\nu^2 C_{ii} - \nu C_{zi} + C_{zz})(\nu^2 C_{jj}- \nu C_{zj} + C_{zz})}} ,
\end{gathered}
\end{equation*}

where $C_{zz}$ is the variance of $Z$, $C_{zi}$ is the covariance between $Z$ and the $i$-th initial vector, and $C_{ij}$ is the covariance between the $i$-th and $j$-th initial vectors.

In the above expression, the covariance terms are constant and may be pre-computed. Thus, each correlation term can be computed in constant-time and the overall number of operations to compute the norm of the correlation matrix is $\mathbf{O}(k^2)$. Note that this value is independent of the size of the data matrix. Further, for typical ensemble sizes, the number of operations for each function evaluation is quite reasonable.

\section{Evaluation}

The method is implemented using Python, NumPy, and SciPy \cite{Numpy,Scipy}. The decision tree estimators are fit using scikit-learn ~\cite{SKLearn}. Source code for the implementation are also made available online \cite{Source}.

\subsection{Experimental Results}

Experiments are performed comparing the shooting regressor to common classification algorithms using the MPG dataset provided by the University of California at Irvine (UCI) \cite{UCIData}. The purpose of this dataset is to predict the average MPG at which a car operates given several properties of the car.

\begin{table}
	\centering
	\begin{tabular}{| l | l | l | l |}
	\hline 
	\multirow{2}{*}{\textbf{Method}} & \multicolumn{2}{|c|}{\textbf{Accuracy}} & \multirow{2}{*}{\textbf{P-Value}} \\
	\cline{2-3}
	 & \textbf{Avg.} & \textbf{Std.} & \\
		\hline
		SR  & 0.8836 & 0.0276 & N/A    \\
		\hline
		GBM & 0.8577 & 0.0401 & 0.0040 \\
		\hline
		RF  & 0.8281 & 0.0286 & 0.0139 \\
		\hline
	\end{tabular}
	\caption{Validation Performance on MPG Dataset}
	\label{table:MPGExpTab}
\end{table}

Three model types are compared: the shooting regressor, random forest, and the gradient boosting machine. Default settings are provided to all models. Specifically, 100 estimators are chosen for all models. The maximum depth of the random forest and SR estimators are unconstrained. The GBM estimators are constrained to a maximum depth of 3. Thirty-two training and validation splits are conducted and each of the three models are fit and evaluated on each respectively. 

\begin{figure}
\center
\includegraphics[width=\columnwidth]{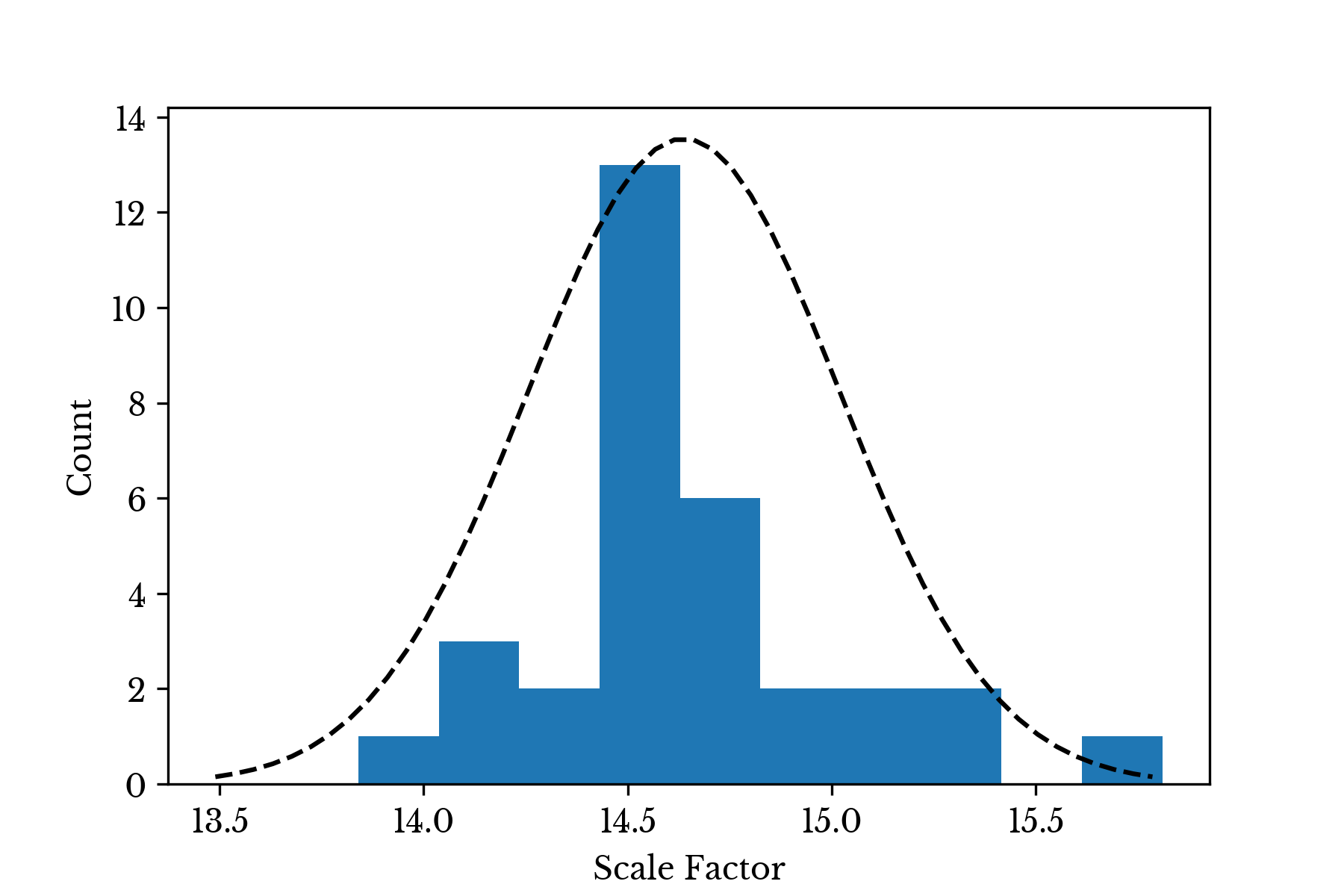}
\caption{Distribution of $\nu$ Values}
\label{fig:SFHist}
\end{figure}

Results from the 32 trials are shown in Table \ref{table:MPGExpTab}. As can be seen, SR performs significantly better than both RF and GBM at the 5\% confidence level. From Figure \ref{fig:SFHist}, it is seen that the value of $\nu$ is relatively consistent over all trials.

\begin{figure}
\center
\includegraphics[width=\columnwidth]{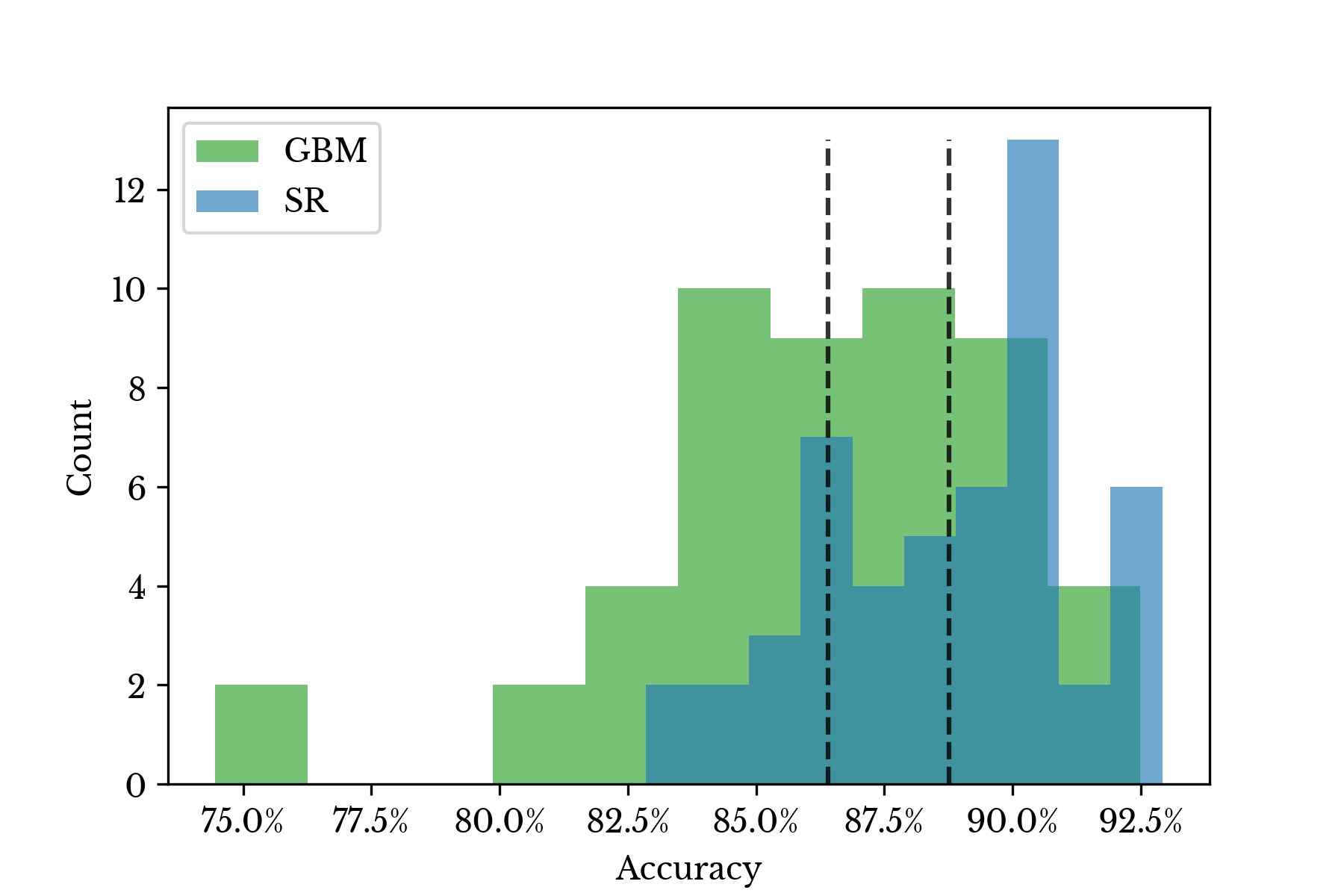}
\caption{Experimental Results for SR and GBM}
\label{fig:SRGBMHist}
\end{figure}

\begin{figure}
\center
\includegraphics[width=\columnwidth]{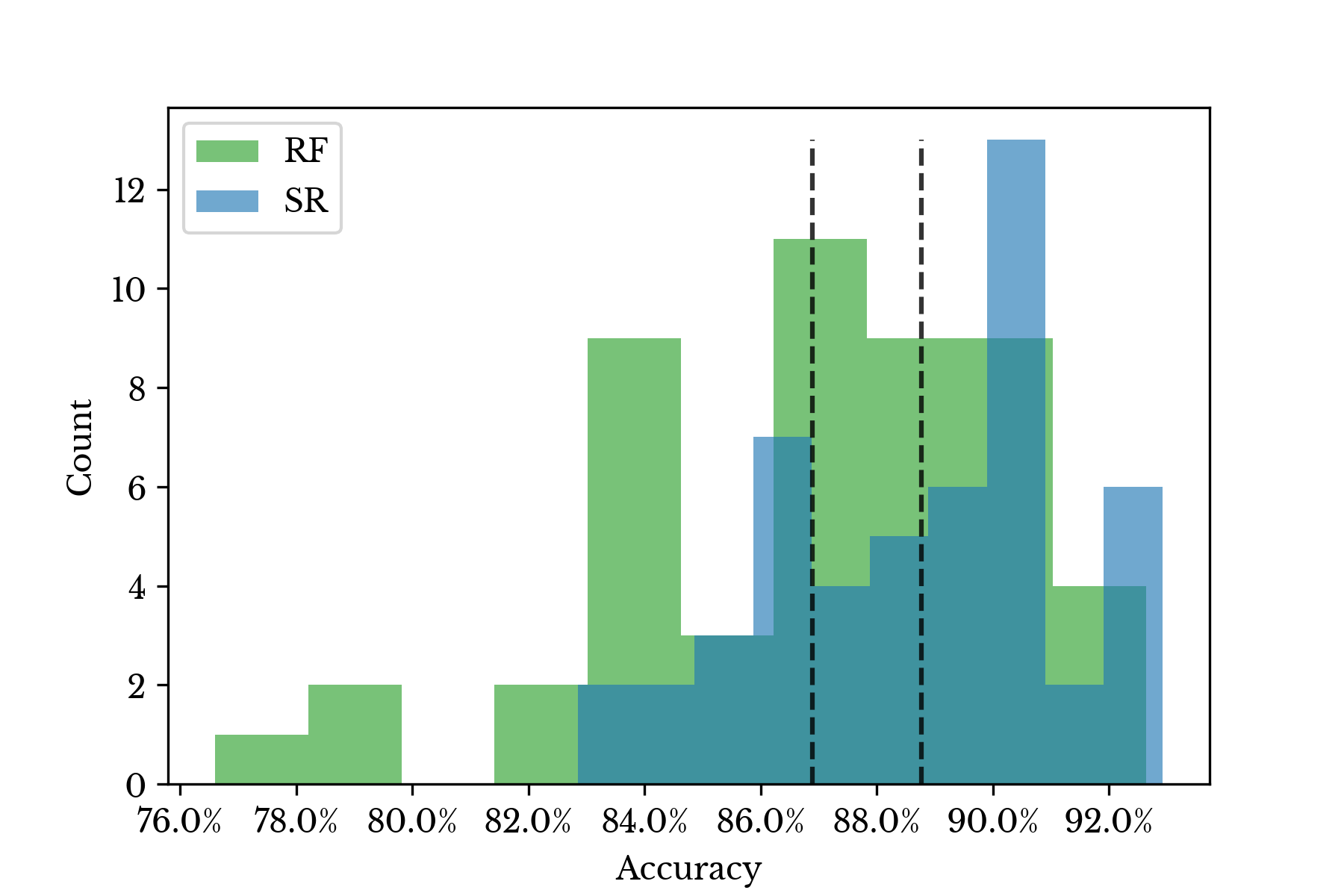}
\caption{Experimental Results for SR and RF}
\label{fig:SRRFHist}
\end{figure}

Histograms are constructed using Matplotlib showing the distributions of the experimental trials \cite{Matplotlib}. The mean values for both model types are shown with dashed black lines. As seen in Figures \ref{fig:SRGBMHist} \& \ref{fig:SRRFHist}, SR outperforms both RF and GBM on this dataset.

\section{Future Work}

Expression \ref{ObjfExp} essentially describes a regularized solution that minimizes correlation. Numerical minimization is presently employed to find a good value of $\nu$ and in practice this approach is efficient when optimized. However, it may be fruitful to explore analytical solutions to Expression \ref{ObjfExp} both for more efficient methods and for further insight into the behavior of the shooting regressor.

\section{Conclusion}

The shooting regressor is an approach to machine learning that incorporates aspects of both random forest and the gradient boosting machine \cite{RandForest,GradBoost}. First, it utilizes the gradient to perform gradient descent though it does so without using a sequential process \cite{GradBoost}. Second, it leverages randomization and a weakly correlated ensemble to achieve a more robust solution. The technique here achieves randomization without the bagging approach employed in random forest \cite{Bagging,RandForest}.

The scaling parameter $\nu$ controls a trade-off between ensemble correlation and precision of the gradient estimation. One of the primary advantages of SR is that good values for $\nu$ can be computed efficiently through numerical optimization. This avoids the need for time-consuming hyper-parameter tuning and provides a potentially more explicit balance between randomization and accuracy than is offered by other methods.


\begin{thebibliography}{99}

\bibitem{Bagging} Breiman, L. (1996). Bagging predictors. \emph{Machine learning, 24}(2), 123-140.

\bibitem{CART} Breiman, L. (2017). \emph{Classification and regression trees.} Routledge.

\bibitem{RandForest} Breiman, L. (2001). Random forests. \emph{Machine learning, 45}(1), 5-32. 

\bibitem{UCIData} Dua, D. and Graff, C. (2019). \href{http://archive.ics.uci.edu/ml}{UCI Machine Learning Repository}. Irvine, CA: University of California, School of Information and Computer Science.

\bibitem{GradBoost} Friedman, J. H. (2001). Greedy function approximation: a gradient boosting machine. \emph{Annals of statistics}, 1189-1232.

\bibitem{AddLR} Friedman, J., Hastie, T., \& Tibshirani, R. (2000). Additive logistic regression: a statistical view of boosting (with discussion and a rejoinder by the authors). \emph{The annals of statistics, 28}(2), 337-407.

\bibitem{ESL} Friedman, J., Hastie, T., \& Tibshirani, R. (2001). The elements of statistical learning (Vol. 1, No. 10). New York: Springer series in statistics.

\bibitem{Matplotlib} Hunter, J. D. (2007). Matplotlib: A 2D graphics environment. \emph{Computing in science \& engineering, 9}(3), 90.

\bibitem{Scipy} Jones, E., Oliphant, T., \& Peterson, P. (2016). SciPy: Open source scientific tools for Python, 2001.

\bibitem{GBGradDesc} Mason, L., Baxter, J., Bartlett, P. L., \& Frean, M. R. (2000). Boosting algorithms as gradient descent. In \emph{Advances in neural information processing systems} (pp. 512-518).

\bibitem{PCA} Pearson, K. (1901). LIII. On lines and planes of closest fit to systems of points in space. The London, Edinburgh, and Dublin Philosophical Magazine and Journal of Science, 2(11), 559-572.

\bibitem{SKLearn} Pedregosa, F., Varoquaux, G., Gramfort, A., Michel, V., Thirion, B., Grisel, O., ... \& Vanderplas, J. (2011). Scikit-learn: Machine learning in Python. \emph{Journal of machine learning research}, 12(Oct), 2825-2830.

\bibitem{Source} Smith, N. ShootingML (2020), \emph{Github repository} \href{https://github.com/nicholastoddsmith/ShootingML}{github.com/nicholastoddsmith/ShootingML}

\bibitem{Numpy} Van Der Walt, S., Colbert, S. C., \& Varoquaux, G. (2011). The NumPy array: a structure for efficient numerical computation. \emph{Computing in Science \& Engineering, 13}(2), 22.

\end{thebibliography}
\end{document}